\documentclass{article} 
\usepackage{single-temp/iclr2025_conference,times}

\usepackage{amsmath,amsfonts,bm}

\def\eqref#1{equation~\ref{#1}}

\def\1{\bm{1}}

\DeclareMathAlphabet{\mathsfit}{\encodingdefault}{\sfdefault}{m}{sl}
\SetMathAlphabet{\mathsfit}{bold}{\encodingdefault}{\sfdefault}{bx}{n}

\pdfoutput=1
\iclrfinaltrue 

\usepackage{multirow}
\usepackage{booktabs}

\usepackage{graphicx}
\usepackage{subcaption}

\usepackage{hyperref}
\usepackage{url}

\usepackage{algorithm}
\usepackage{algorithmicx}
\usepackage{algpseudocode}
\usepackage{amsmath}

\usepackage{csquotes}
\newcommand{\myFrame}{ReSpec}

\title{When, What, and How: Rethinking Retrieval-Enhanced Speculative Decoding}

\author{Min Fang, Zhihui Fu, Qibin Zhao \& Jun Wang \thanks{Corresponding author.} \\
OPPO Research Institute, Shenzhen, China \\
\texttt{mfang.cs@gmail.com, \{luca,zhaoqibin\}@oppo.com, junwang.lu@gmail.com} 
}

\usepackage{enumitem}

\begin{document}

\maketitle

\begin{abstract}
Speculative decoding (SD) has emerged as an effective technique to accelerate large language model (LLM) inference without compromising output quality.
However, the achievable speedup largely depends on the effectiveness of the drafting model.
While model-based methods like EAGLE-2 are accurate but costly, retrieval-enhanced methods like SAM-Decoding rely on heuristic switching strategies that often trigger unnecessary retrievals.
To address this, we propose \textbf{\myFrame}~(\textbf{Re}trieval-enhanced \textbf{Spe}culative Decoding), a novel framework that transforms heuristic drafter switching into adaptive decision-making. 
\myFrame~features three core innovations: 1) An \textbf{entropy-guided adaptive trigger} quantifies contextual predictability to initiate retrieval only when uncertainty is low, avoiding costly low-quality speculations. 2) A \textbf{feedback-driven candidate selection} leverages historical feedback to organize multiple high-quality candidates for parallel verification, maximizing retrieval utility. 3) A source-aware \textbf{relaxed verification strategy} applies strict checks to model-generated drafts while using a relaxed verification for retrieved drafts, achieving a better balance between accuracy and efficiency. 
Extensive experiments on Spec-Bench demonstrate that \myFrame~achieves state-of-the-art acceleration,
outperforming EAGLE-2 and SAM-Decoding by over $33\%$ and $25\%$, respectively, while maintaining output quality.
\end{abstract}
\section{Introduction}

Large Language Models (LLMs) have achieved remarkable success across diverse domains~\citep{gpt4,llama2}, yet their deployment remains constrained by substantial inference latency~\citep{pope2023efficiently,kwon2023efficient}. Speculative Decoding (SD) alleviates this issue by using a lightweight draft model to propose candidate tokens, which are then verified by the target LLM in parallel~\citep{spd3,spd,spd1,spd2}.
However, significant divergence between draft and target model distributions can lead to low acceptance rates, limiting potential speedup~\citep{survey1,survey2}.
This makes the design of effective drafting strategies critical, a challenge we term the \textit{Drafter's Dilemma}.

On one hand, \textbf{model-based drafters}, such as SPD~\citep{spd} and EAGLE series~\citep{eagle,eagle2,eagle3}, generate drafts using a smaller model or auxiliary modules trained on the target LLM. 
While they improve acceptance rates, they require careful system integration or substantial training costs.
On the other head, model-free \textbf{retrieval-based drafters}, such as CopySpec~\citep{copyspec} and Token Recycling~\citep{tokenrecycle}, offer training-free and lightweight alternatives but suffer from limited retrieval quality.
Recent work has explored \textbf{hybrid} strategies that combine model-based drafting with retrieval.
For example, RASD~\citep{rasd} enhances draft generation by fusing retrieval results with model-based draft, while SAM-Decoding~\citep{samd} uses a Suffix Automaton (SAM) for context retrieval with dynamic fallback to auxiliary drafters like EAGLE~\citep{eagle,eagle2,eagle3} when no suitable match is found.
These efforts show particular promise in tasks with high text repetition, such as document summarization and retrieval-augmented generation (RAG)~\citep{copyspec,rest,suffixdecoding}. 

However, current retrieval-enhanced speculative decoding methods still face three key limitations:


\textbf{(1) Context-Ignorant Retrieval Triggering:} Retrieval is typically triggered by fixed prefix matching rules, either single~\citep{samd,suffixdecoding} or multi-token~\citep{copyspec}, ignoring contextual predictability. This causes unnecessary atempts in uncertain scenarios where retrieval offers little benefit, wasting computation on failed verification and preventing adaptive trade-offs between lightweight retrieval and robust model-based drafting.


\textbf{(2) Greedy Candidate Selection:} Existing methods retrieve only from the earliest matched position~\citep{samd,copyspec}, overlooking potentially better continuations from later matches. Simply considering all match positions would incur excessive verification costs on low-quality candidates, yet without a principled selection strategy, many high-quality options are discarded.
Moreover, once retrieval is triggered, they always construct and verify candidates regardless of quality, wasting computation on unpromising drafts and limiting retrieval effectiveness.

\textbf{(3) Rigid Verification Policy:} Current works enforce strict token-level alignment with target predictions, ensuring lossless generation but rejecting semantically valid drafts~\citep{judgedecoding}. This issue is amplified in retrieval-based drafting, where drafts come from historical text rather than the model's distribution~\citep{wang2025alignment}. Thus, contextually appropriate drafts are discarded for minor deviations, leading to low acceptance rates that negate retrieval benefits.

These limitations highlight a critical gap: existing methods lack adaptive mechanisms for informed retrieval decisions. We address this with \textbf{\myFrame}~(\textbf{Re}trieval-enhanced \textbf{Spe}culative Decoding), which transforms retrieval-enhanced speculative decoding from heuristic-driven to principled and adaptive.
Our key insight is that effective hybrid decoding requires intelligent orchestration of draft sources based on contextual signals and historical feedback, rather than simple combination. 
specifically, \myFrame~introduces three innovations to address the identified limitations:
\begin{itemize}[leftmargin=*, align=left]


    \item \textbf{Entropy-Guided Adaptive Trigger:} We use information entropy as a principled measure of predictability, building on recent findings linking entropy to generation quality and reasoning reliability~\citep{entropyselect,entropyselect1}. By averaging entropy across retrieval prefixes of varying lengths, we detect high-confidence contexts where low entropy indicates predictable patterns suitable for retrieval. Retrieval is triggered only when entropy falls below a learned threshold, avoiding unnecessary retrieval attempts and ensuring efficient resource use.

    \item \textbf{Feedback-Driven Candidate Selection:} We employ exponential moving average (EMA) scoring to track retrieval position quality dynamically~\citep{ema}. Each match position maintains a score reflecting its historical performance while preserving flexibility to explore new candidates. 
    Only top-ranked positions with scores above a learned threshold are organized into a draft tree for verification; otherwise, decoding directly fallbacks to the model-based drafter. This design maximizes retrieval benefits while avoiding exhaustive verification costs.

    \item \textbf{Relaxed Verification Strategy:} We apply source-specific verification policies tailored to draft origin. Model-based drafts undergo strict lossless verification to preserve distributional consistency~\citep{spd}, while retrieval-based drafts employ relaxed verification inspired by Spec-Verification~\citep{Spec-verification} and enhanced with a look-ahead tolerance mechanism. 
    By aligning verification criteria with draft origin, we optimize the trade-off between speedup and quality in retrieval-intensive scenarios.

\end{itemize}

Figure~\ref{fig:overview} gives three cascading limitations via a document summarization example and shows \myFrame's systematic solutions. Our solution contains: entropy-guided triggering for well-motivated retrieval attempts, feedback-driven selection identifies promising candidates, and source-aware relaxed verification maximizes draft acceptance while preserving fidelity.
Our main contributions are:



\begin{itemize}[leftmargin=*, align=left]
    \item We identify and formalize three critical limitations in current retrieval-enhanced speculative decoding methods that prevent effective hybrid integration.
    
    \item We propose \myFrame, a principled framework featuring entropy-guided triggering, feedback-driven selection, and source-aware relaxed verification that transforms retrieval-enhanced from a heuristic add-on to an intelligent and adaptive design.
    
    \item We achieve state-of-the-art acceleration results on diverse benchmarks, demonstrating the effectiveness of adaptive and context-aware control in hybrid inference systems.
    
\end{itemize}
\section{Related Work}
\subsection{Drafter Design in Speculative Decoding}
Speculative decoding can be broadly categorized by the nature of the drafting component, ranging from \emph{model-based} to \emph{model-free} approaches, and further to hybrid variants combining the two.

\textbf{Model-based Methods.}
The field of speculative decoding has evolved from using standalone, smaller LMs as drafters \citep{spd} to more integrated ones. Medusa \citep{medusa} introduced the concept of training multiple decoding heads on the target LLM itself, enabling parallel draft generation without a separate model.
Based on this, some related semi-autoregressive variants~\citep{lin2025bita} or non-autoregressive paradigms~\citep{diffuspec} have been proposed to enhance the overall decoding speed.
EAGLE series \citep{eagle,eagle2,eagle3} and its variants~\citep{eagle-v1,eagle-v2} advanced this direction by shifting from token-level to feature-level prediction. EAGLE \citep{eagle} predicted hidden states rather than discrete tokens, EAGLE-2 \citep{eagle2} introduced context-aware draft trees, and EAGLE-3 \citep{eagle3} leveraged multi-layer feature fusion for direct token prediction.
However, these methods typically require extensive model-specific training.

\textbf{Model-free Methods.}
To reduce the overhead of auxiliary models, some model-free methods has emerged under text repetition, i.e., retrieval-based. 
Prompt Lookup Decoding (PLD) \citep{pld} was an early example, retrieving n-grams directly from the input prompt. REST \citep{rest} expands this by retrieving from a large external text corpus using a suffix array. CopySpec \citep{copyspec} uses a hash map to find and copy repeated sequences from the generation history. 
The efficiency of these retrieval mechanisms is paramount, leading to the adoption of advanced data structures like suffix trees and suffix automata for highly efficient longest-suffix matching \citep{suffixdecoding, samd}. 
Another variant, Token Recycling \citep{tokenrecycle}, cleverly re-uses top-$k$ candidates from previous steps, storing them in an adjacency matrix for future drafting. 

\textbf{Hybrid and Augmented Methods.}
Recognizing the complementary strengths of model-based and retrieval-based methods, hybrid approaches have been proposed.
RASD \citep{rasd} augment a generative draft tree by using retrieval results from the context to prune and fuse it. 
SAM-Decoding \citep{samd}, the direct predecessor to our work, uses a suffix automaton for efficient retrieval and falls back to model-based one (e.g., EAGLE-2 or Token Recycling) when no match is found. 
CopySpec~\citep{copyspec} also supports heuristic drafter switching like SAM-Decoding.
Although promising, existing hybrids often rely on heuristic switching rules. Our work builds on this line by proposing an adaptive, information-theoretic criterion for drafter selection.

\subsection{Relaxed and Approximate Verification}
The standard SD framework guarantees output distributions identical to the target model \citep{spd}, a property known as being \textit{lossless}. However, the pursuit of efficient speculative decoding has led to significant innovations in relaxed verification mechanisms, progressively moving beyond strict distributional alignment. 
Early work by Medusa \citep{medusa} introduced typical acceptance criteria, trading exact probability matching for entropy-based thresholds to achieve speedups. This direction was advanced by SPRINTER \citep{SPRINTER}, which introduced a small, trained verifier to \textit{predict} acceptance, avoiding frequent calls to the target LLM at the cost of approximation. 
Most notably, Judge Decoding \citep{judgedecoding} trains a linear \textit{judge} head on the target model's embeddings to assess the quality of a draft token, rather than its strict probabilistic alignment.
BiLD \citep{BiLD} uses a \textit{rollback} policy based on the distance between the small and large models' probability distributions. 
Spec-Verification \citep{Spec-verification} consolidated these advances through a training-free heuristic combining top-ranked sampling and adaptive logit thresholds, delivering significant performance improvements while ensuring output quality. 
Collectively, these works reveal an evolution from rigid distribution-matching toward adaptive, semantics-aware verification that balances computational efficiency with controlled quality tradeoffs.
Similarly, several works have explored applying speculative decoding and relaxed acceptance conditions to visual auto-regressive or vision-language-action models to improve their generation speed~\citep{vlmrelaxed1,vlmrelaxed2}.
Our relaxed verification is inspired by this direction but is intentionally designed to be a lightweight, training-free heuristic, making it more practical to deploy.
\section{Methodology}
\begin{figure}[t]
    \centering
    \includegraphics[width=\linewidth]{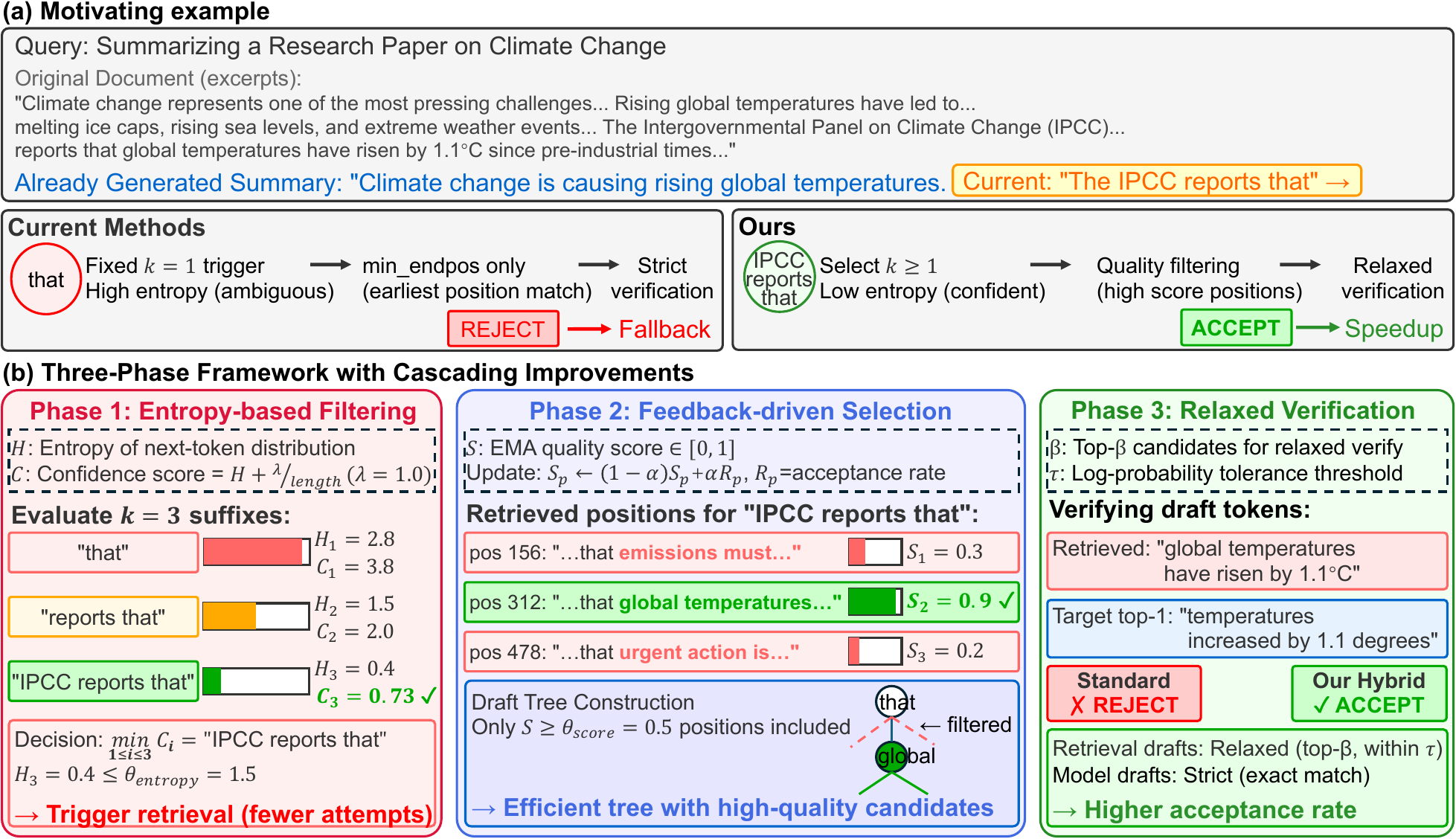}
    \caption{Overview of three key limitations in current hybrid speculative decoding methods and our proposed solutions. 
    \textbf{(a)} Motivating example from document summarization showing how current methods fail with ambiguous triggers while our framework succeeds through entropy-guided selection. 
    \textbf{(b)} Detailed workflow of our three-phase approach. Phase 1 evaluates $k=3$ suffix candidates using confidence score $C = H + \lambda/\textit{length}$ where $H$ is the entropy of next-token distribution. Phase 2 filters retrieved positions based on EMA quality scores $S_p$ updated via $S_p \leftarrow (1-\alpha)S_p + \alpha R_p$ where $R_p$ is the historical acceptance rate. Phase 3 applies hybrid verification: relaxed (top-$\beta$ with tolerance $\tau$) for retrieval drafts and strict for model-generated drafts. 
    }
    \label{fig:overview}
\end{figure}

\subsection{Overview of \myFrame}


\myFrame~operates as a hybrid system that dynamically selects between two complementary drafting modules at each generation step: a lightweight \textit{Retriever Module} for predictable content and a powerful \textit{Drafter Module} for complex generation. The decision is governed by an entropy-guided adaptive trigger that assesses the predictability of the current context. When the trigger identifies high-confidence copying opportunities, it activates the \textit{Retriever Module} and applies hybrid verification to its outputs. Otherwise, the system defaults to the \textit{Drafter Module} with standard strict verification. This design enables efficient exploitation of repetitive patterns while maintaining generation quality for novel content.
Figure~\ref{fig:overview} illustrates the design of \myFrame~through a detailed example, highlighting its three-phase framework with cascading improvements.

\subsection{Retriever Module: Entropy-Guided Adaptive Drafting}

The \textit{Retriever Module} addresses fundamental limitations of naive retrieval mechanisms through two key innovations: an entropy-based selection trigger and a feedback-driven candidate scoring process.

\subsubsection{Entropy-Guided Adaptive Trigger}
The core of our adaptive trigger is the principle that a retrieval attempt is only worthwhile if the immediate future is highly predictable. We measure this predictability by the mean Shannon entropy of the target model's output distribution at the previous step, a metric recently validated as a reliable signal for reasoning trace quality \citep{entropyselect,entropyselect1}.

Given an inference sequence $X_{1:t}$ comprising both prompt and generated tokens, let $P_t$ denote the target model's probability distribution for the next token $X_{t+1}$. For a suffix sequence $X_{t-k+1:t}$ serving as the retrieval key (a contiguous token sequence of length $k$ for indentifying relevant contextual patterns), we define its average predictive entropy as:
\begin{equation}
H(X_{t-k+1:t}) = \frac{1}{k} \sum_{i=1}^{k} \left( -\sum_{v \in \mathcal{V}} P_{t-k+i}(v) \log P_{t-k+i}(v) \right)
\end{equation}
where $\mathcal{V}$ represents the model's vocabulary.

At each generation step, our algorithm performs the following:
\begin{enumerate}[leftmargin=*, align=left]
    \item \textbf{Entropy Computation}: For each suffix length $k \in \{1, \ldots, l\}$ (where $l$ is the maximum lookback length), we extract the corresponding suffix $X_{t-k+1:t}$ and compute its average entropy $H_k=H(X_{t-k+1:t})$.
    
    \item \textbf{Confidence Scoring}: We calculate a confidence score that balances entropy with length preference:
    \begin{equation}
    C_k = H_k + \frac{\lambda_e}{k}
    \end{equation}
    where $\lambda_e$ controls the entropy-length trade-off.
    
    \item \textbf{Key Selection}: We identify the optimal suffix:
    \begin{equation}
    k^* = \arg\min_{k \in \{1,\ldots,l\}} C_k
    \end{equation}
    with corresponding minimum entropy $H_{\text{min}} = H_{k^*}$.
    
    \item \textbf{Trigger Decision}: Retrieval is triggered if $H_{\text{min}} \leq \theta_{\text{entropy}}$ ($\theta_{\text{entropy}}=1.5$ is used in our experiments \S\ref{sec:experiments}), indicating sufficient predictability. Otherwise, control passes to the \textit{Drafter Module}.
\end{enumerate}

This mechanism ensures that we only attempt to retrieve from context when the model demonstrates high confidence (low uncertainty) in its recent predictions, indicating a stable and predictable generation trajectory that is suitable for speculative retrieval.

\subsection{Feedback-Driven Candidate Selection}

We introduce a feedback-driven selection mechanism that learns from past retrieval outcomes to identify high-quality match positions. This approach overcomes the limitations of both greedy selection and exhaustive verification through dynamic scoring and intelligent candidate curation.

\textbf{Dynamic Candidate Scoring with EMA.}
We maintain a quality score $S_{\text{pos}}$ for each historical match position, which is updated at every retrieval attempt using an Exponential Moving Average (EMA):
\begin{equation}
    S_{t+1} = (1 - \alpha) \cdot S_t + \alpha \cdot R_t
    \label{eq:ema_update}
\end{equation}
where $S_t$ is the score of the position before the update, and $S_{t+1}$ is the new score. $\alpha$ is the update rate, which controls the sensitivity to recent outcomes. $R_t$ represents the instantaneous performance, calculated as the acceptance rate of the draft generated from this position:
\begin{equation}
    R_t = \frac{L_{\text{accepted}}}{L_{\text{draft}}}
    \label{eq:rate_calc}
\end{equation}
Here, $L_{\text{accepted}}$ is the number of tokens accepted from the draft, and $L_{\text{draft}}$ is its total length. This formulation inherently penalizes poor suggestions: if none of the draft tokens are accepted (i.e., $L_{\text{accepted}} = 0$), then $R_t=0$, causing its score $S_t$ to decay. Newly encountered positions are initialized with a neutral score $S_0$ (e.g., 0.5), ensuring exploration of unseen candidates.


\textbf{Selection Algorithm.}
Leveraging the dynamic scores, we employ a hybrid selection algorithm that combines quality-based filtering with a fixed-size candidate pool to balance performance and computational stability. The algorithm proceeds in the following steps:
\begin{enumerate}[leftmargin=*, align=left]
    \item \textbf{Retrieval and Scoring}: Retrieve all match positions from SAM based on the given retrieval key and collect their current quality scores.
    
    \item \textbf{Threshold Filtering}: Eliminate candidates with scores below threshold $\theta_{score}$, preventing verification of consistently low-quality continuations.
    
    \item \textbf{Ranking and Selection}: Select top-3 (fixed in our experiments \S\ref{sec:experiments}) candidates by score for parallel verification.
\end{enumerate}

\textbf{Comprehensive Feedback Loop.}
Post-verification, we update all match positions scores based on observed performance $R_t$. The accepted candidate updates with its true acceptance rate, while rejected candidates receive $R_t = 0$. This balanced feedback enables efficient exploration-exploitation trade-offs and rapid convergence to reliable historical continuations.

\subsection{Source-Aware Relaxed Verification}
To enhance the utilization of drafting results, \myFrame~employs a conditional verification strategy based on the source of the draft tokens.

\textbf{For Drafter Module outputs:} We apply standard lossless speculative sampling to maintain strict distributional consistency with the target model \citep{spd}.

\textbf{For Retriever Module outputs:} We propose \textbf{Relaxed Verification with Look-ahead Tolerance}, an extension of prior work on flexible verification Spec-Verification \citep{Spec-verification}, specifically tailored for retrieved drafts. For a candidate retrieved token $\tilde{x}_{t+1}$ that is rejected by the target model, we apply a relaxed check. In particular, our verification accepts the token if the following two conditions, similar to those in Spec-Verification, are satisfied:
\begin{gather}
\tilde{x}_{t+1} \in \operatorname{TopK}_k(p_t), \label{eq:constraint1}\\
\log p_t(x^{(1)}_{t+1}) - \log p_t(\tilde{x}_{t+1}) \leq \tau, \label{eq:constraint2}
\end{gather}
where $x^{(1)}_{t+1} = \arg\max_{v} p_t(v)$ is the greedy top-1 token, and $\tau$ is a tolerance margin.

Beyond token-wise \textbf{Relaxed Verification}, we introduce a \textbf{Look-ahead Tolerance} mechanism that integrates this relaxed verifying with standard greedy verification.
Unlike Spec-Verification, which applies relaxed checks to every token, we selectively apply them a fixed number of times to balance accuracy and computational efficiency.
For a candidate sequence $\tilde{x}_{t+1:t+L}$, we:
\begin{enumerate}[leftmargin=*, align=left]
    \item Establish the initial greedy-match prefix length $\ell_0$
    \item Maintain tracking variables: $\textit{current} \gets \ell_0$, $\textit{last} \gets \ell_0$
    \item Perform up to $\varPhi$ relaxed attempts, each:
    \begin{itemize}
        \item Apply relaxed verification at position $\textit{current}$
        \item If successful and followed by $\xi \geq m$ look-ahend tolerance greedy matches, extend acceptance and update $\textit{last} \gets s + \xi - 1$, $\textit{current} \gets \textit{last}$
        \item Otherwise, advance to the next position and increment $\textit{current} \gets \textit{current} + 1$
    \end{itemize}
\end{enumerate}

The procedure ends when either the loop completes $\varPhi$ attempts or it is terminated early by a relaxed-verify failure. The final returned accepted length is the value of $last$, which represents the longest path found that satisfies all the verification and tolerance conditions. 
This rule accepts tokens that are semantically plausible and highly probable, even if they are not the single most likely token. This is particularly useful for copied text, where functionally equivalent synonyms (e.g., \enquote{the}, \enquote{that}) might otherwise cause a rejection. This provides a robust and lightweight alternative to methods like Judge Decoding \citep{judgedecoding}, specifically tailored for verifying retrieved drafts while maintaining controlled accuracy.

\subsection{Putting It All Together}

Algorithm~\ref{alg:myframe} presents the core workflow \myFrame, which operates through three coordinated steps:

\textbf{Step 1: Entropy-Guided Triggering} (Lines 2-9): The algorithm evaluates suffix candidates of varying lengths to identify the optimal retrieval key. For each suffix, it computes a confidence score that balances predictive entropy against length preference. Retrieval proceeds only when both entropy and existence conditions are satisfied.

\textbf{Step 2: Adaptive Retrieval and Verification} (Lines 10-23): Upon successful triggering, the system retrieves candidate continuations and applies EMA-based filtering to select high-quality positions. These candidates undergo relaxed verification with look-ahead tolerance. Following verification, the EMA scores are updated based on observed acceptance rates, creating a feedback loop that improves future selection quality.

\textbf{Step 3: Fallback Generation} (Lines 24-27): When retrieval conditions are not met, the system seamlessly defaults to model-based drafting with strict verification, ensuring robust generation across all contexts.

Finally, the accepted tokens $X_{\text{accepted}}$ are appended to the context, used to update SAM, and returned (Lines 28-30). This adaptive design ensures efficient retrieval for predictable contexts while maintaining generation quality through principled verification strategies.

\begin{algorithm}[hb]
\caption{Adaptive Retrieval-Enhanced Speculative Decoding Framework}
\label{alg:myframe}
\begin{algorithmic}[1]
\Require Target model $M_T$; Draft model $M_D$; Suffix Automaton $\mathcal{SAM}$; Current context $X$; Maximum suffix length $l$; Entropy threshold $\theta_{\text{entropy}}$; Length penalty $\lambda_e$; EMA update rate $\alpha$; Position quality threshold $\theta_{\text{score}}$.
\Ensure Accepted tokens $X_{\text{accepted}}$
\Statex
\Comment{Step 1: Entropy-guided adaptive triggering}
\State Initialize $C_{\min} \gets \infty$, $H_{\text{best}} \gets \infty$, $\text{key}^* \gets \text{None}$
\For{$k = 1$ to $l$}
    \State $\text{key}_k \gets$ suffix of $X$ with length $k$
    \State $H_k \gets \textsc{ComputeMeanEntropy}(M_T, \text{key}_k)$
    \State $C_k \gets H_k + \lambda_e / k$ \Comment{Balance entropy and length}
    \If{$C_k < C_{\min}$}
        \State $C_{\min} \gets C_k$, $H_{\text{best}} \gets H_k$, $\text{key}^* \gets \text{key}_k$
    \EndIf
\EndFor
\Statex
\If{$H_{\text{best}} \leq \theta_{\text{entropy}}$ \textbf{and} $\mathcal{SAM}.\textsc{Contains}(\text{key}^*)$}
    \Comment{Step 2: Retrieval with feedback-driven selection}
    \State $\mathcal{P} \gets \mathcal{SAM}.\textsc{RetrievePositions}(\text{key}^*)$
    \State $\mathcal{P}_{\text{filtered}} \gets \{p \in \mathcal{P} : S_p \geq \theta_{\text{score}}\}$ \Comment{Quality filtering}
    \State $\mathcal{P}_{\text{selected}} \gets \textsc{SelectTopK}(\mathcal{P}_{\text{filtered}}, k=3)$ \Comment{Select best candidates}
    \State $T_{\text{draft}} \gets \textsc{BuildDraftTree}(\mathcal{P}_{\text{selected}})$
    \State $X_{\text{accepted}}, p_{\text{winner}} \gets \textsc{RelaxedVerify}(M_T, T_{\text{draft}})$
    \Statex
    \Comment{Update EMA scores based on performance}
    \For{each $p \in \mathcal{P}_{\text{selected}}$}
        \If{$p = p_{\text{winner}}$}
            \State $R_p \gets L_{\text{accepted}} / L_{\text{draft}}(p)$
        \Else
            \State $R_p \gets 0$ \Comment{Penalize non-selected candidates}
        \EndIf
        \State $S_p \gets (1-\alpha) \cdot S_p + \alpha \cdot R_p$ \Comment{EMA update}
    \EndFor
\Else
    \Comment{Step 3: Fallback to model-based drafting}
    \State $T_{\text{draft}} \gets M_D.\textsc{Generate}(X)$
    \State $X_{\text{accepted}} \gets \textsc{StrictVerify}(M_T, T_{\text{draft}})$
\EndIf
\Statex
\State $X \gets X \oplus X_{\text{accepted}}$ \Comment{Append accepted tokens}
\State $\mathcal{SAM}.\textsc{Update}(X_{\text{accepted}})$ \Comment{Update suffix automaton SAM}
\State \Return $X_{\text{accepted}}$
\end{algorithmic}
\end{algorithm}
\section{Experiments}
\label{sec:experiments}

This section presents a thorough empirical evaluation of \myFrame~across diverse task domains. Detailed results demonstrate its effectiveness against strong baselines, and ablation studies highlight the contributions of its core components.

\subsection{Experimental Setup}
\textbf{Models and Benchmark.} 
To demonstrate the scalability and general applicability of \myFrame, we evaluate it  across three widely-used model families: Vicuna-7B-v1.3 \citep{vicuna}, Llama-3.1-8B-Instruct \citep{llama3}, and Qwen2-7B-Instruct \citep{qwen}. 
Our primary evaluation is conducted on Spec-Bench \citep{spec-bench}, a comprehensive benchmark designed for speculative decoding that includes six diverse tasks: Multi-turn Conversation (MT), Translation (Trans), Summarization (Sum), Question Answering (QA), Mathematical Reasoning (Math), and Retrieval-Augmented Generation (RAG), derived from datasets including MT-Bench \citep{mt-bench}, WMT14 DE-EN, CNN/Daily Mail \citep{sum}, Natural Questions \citep{qa}, GSM8K \citep{gsm8k}, and DPR \citep{rag}.

\textbf{Baselines.} We compare \myFrame~against a set of strong baselines from different categories:  
\begin{itemize}[leftmargin=*, align=left]
    \item \textbf{EAGLE-2} \citep{eagle2}: A model-based self-speculative decoding framework, serving as the powerful auxiliary drafter in our hybrid system.
    \item \textbf{SAM-Decoding} \citep{samd}: A retrieval-enhanced speculative decoding framework that leverages suffix automata for retrieval and incorporates EAGLE-2 as an auxiliary drafter.
\end{itemize}

\textbf{Metrics.} We use the following aspects to comprehensively evaluate speculative decoding methods:  
\begin{itemize}[leftmargin=*, align=left]
    \item \textbf{Speedup Ratio:} The relative wall-clock speedup compared to standard autoregressive decoding.
    \item \textbf{Mean Accepted Tokens (MAT):} The average number of tokens accepted in each generation step.
    \item \textbf{Throughput:} The average number of tokens generated per second.
    \item \textbf{Accuracy:} The average GPT-4o score (on a 10-point scale) assigned to generated outputs, where higher scores indicate better alignment and correctness \citep{mt-bench}.
\end{itemize}

\textbf{Implementation Details.}
All experiments are conducted on a server equipped with an 11-core CPU and an NVIDIA A100 GPU (80GB), using PyTorch 2.3.0, Transformers 4.51.3, and CUDA 12.1. 
We adopt greedy decoding for strict verification and employ FP16 precision with a batch size of 1. 
Both the baseline SAM-Decoding and our method did not use any external corpus, and all retrieval was based solely on the input prompt and the generated context corpus.
We follow the automaton construction and retrieval procedures described in \citep{samd}, with the default draft size set to 60 and the maximum lookback length to 3.
Besides, for all baselines and our model-based drafter, the default configurations from their respective original papers were adopted.




\subsection{Performance on Spec-Bench}
\textbf{Overall Performance Comparison.} 
Table~\ref{tab:mat} summarizes the main results, comparing the performance of \myFrame~against strong baselines, EAGLE-2 and SAM-Decoding, on the Vicuna-7B and Qwen2-7B models. The results demonstrate that \myFrame~achieves comprehensive and consistent superiority across all measured performance indicators.

In terms of overall efficiency on Vicuna-7B, \myFrame~attains a 3.05x speedup and a throughput of 133.10 tokens/s. This is a significant improvement over both EAGLE-2 (2.30x speedup, 100.24 tokens/s) and SAM-Decoding (2.45x speedup, 106.74 tokens/s).
Furthermore, this leadership extends to all task-specific benchmarks. \myFrame~achieves the competitive accept length in every individual task (MT, Sum, etc.), indicating that its robust across different application scenarios, from Math to RAG. This across-the-board improvement highlights the effectiveness of our approach.

The superiority of \myFrame~is further validated on the Qwen2-7B, where it again leads with a 2.62x speedup. This consistent performance across different model architectures underscores the robustness and generalizability of our method as a universally effective solution for accelerating inference.

\begin{table}[t]
\caption{Overall efficiency and task-specific acceptance length on the Spec-Bench benchmark across different model families. \myFrame~consistently delivers the highest inference efficiency.}
\label{tab:mat}
\centering
\resizebox{\linewidth}{!}{
\begin{tabular}{@{}ccccccccccc@{}}
\toprule
Model                         & Method           & MT            & Trans         & Sum           & QA            & Math          & RAG           & MAT           & Tokens/s        & Speedup       \\ \midrule
\multirow{3}{*}{Vicuna-7B}    & EAGLE-2          & 4.78          & 3.32          & 3.98          & 3.73          & 4.74          & 4.00          & 4.36          & 100.24          & 2.30x          \\
                              & SAM-Decoding     & 5.00          & 3.29          & 4.74          & 4.06          & 4.59          & 4.27          & 4.63          & 106.74          & 2.45x          \\ 
                              & \textbf{\myFrame} & \textbf{5.28} & \textbf{3.32} & \textbf{7.63} & \textbf{4.04} & \textbf{4.98} & \textbf{5.05} & \textbf{5.26} & \textbf{133.10} & \textbf{3.05x} \\ \midrule
\multirow{3}{*}{Qwen2-7B}     & EAGLE-2          & 3.93          & 3.06          & 3.64          & 3.19          & 4.37          & 3.73          & 3.80          & 97.09           & 2.27x          \\
                              & SAM-Decoding     & 3.97          & 3.07          & 3.79          & 3.21          & 4.19          & 3.79          & 3.83          & 87.27           & 2.04x          \\
                              & \textbf{\myFrame} & \textbf{4.06} & \textbf{3.06} & \textbf{5.03} & \textbf{3.21} & \textbf{4.37} & \textbf{3.93} & \textbf{4.07} & \textbf{111.89} & \textbf{2.62x} \\ \midrule
\end{tabular}}
\end{table}

\begin{figure}[!ht]
    \centering
    \begin{subfigure}[b]{0.43\textwidth}
        \includegraphics[width=\linewidth]{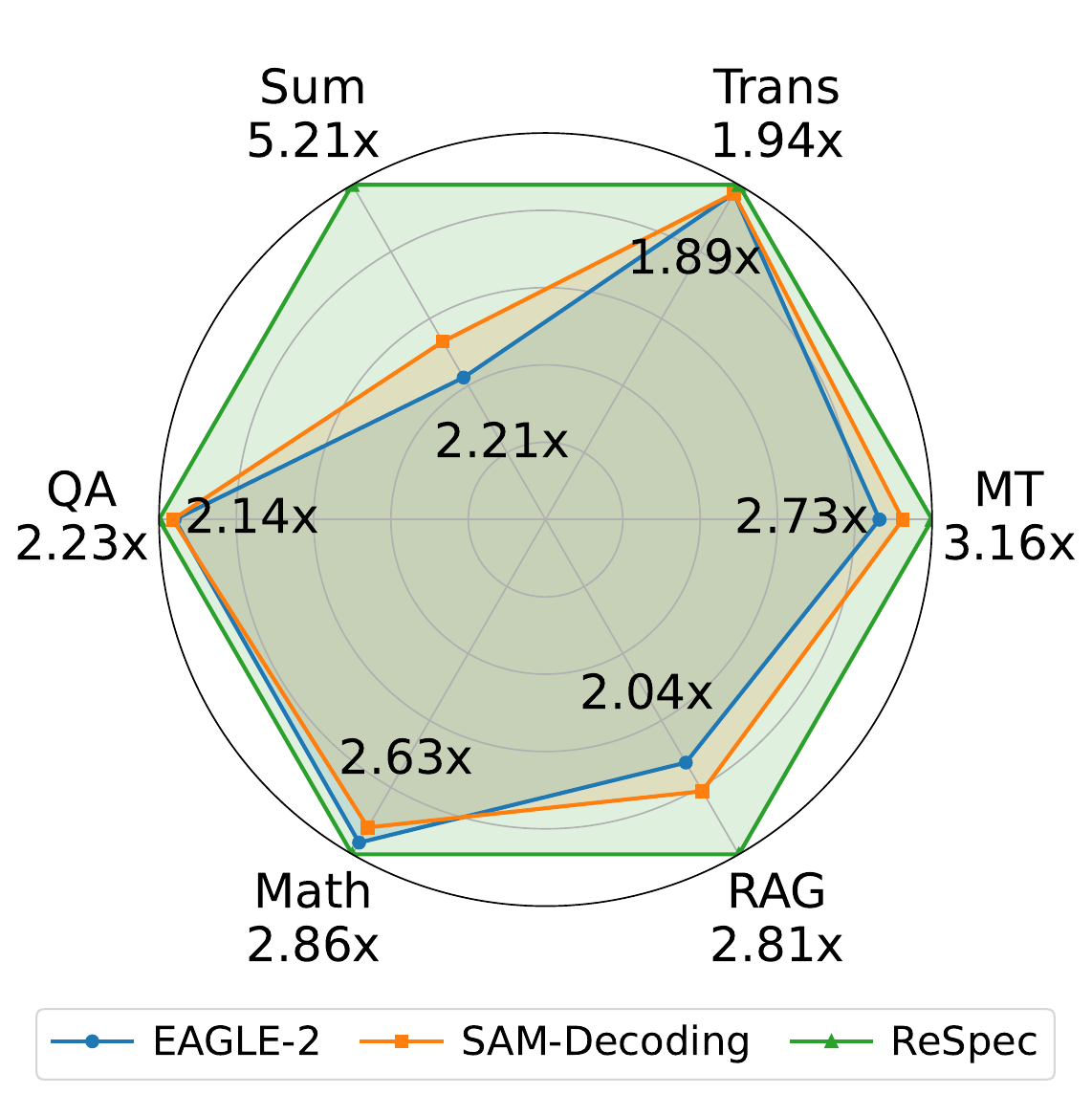}
        \caption{Vicuna-7B-v13}
        \label{fig:sub1}
    \end{subfigure}
    \hfill
    \begin{subfigure}[b]{0.43\textwidth}
        \includegraphics[width=\linewidth]{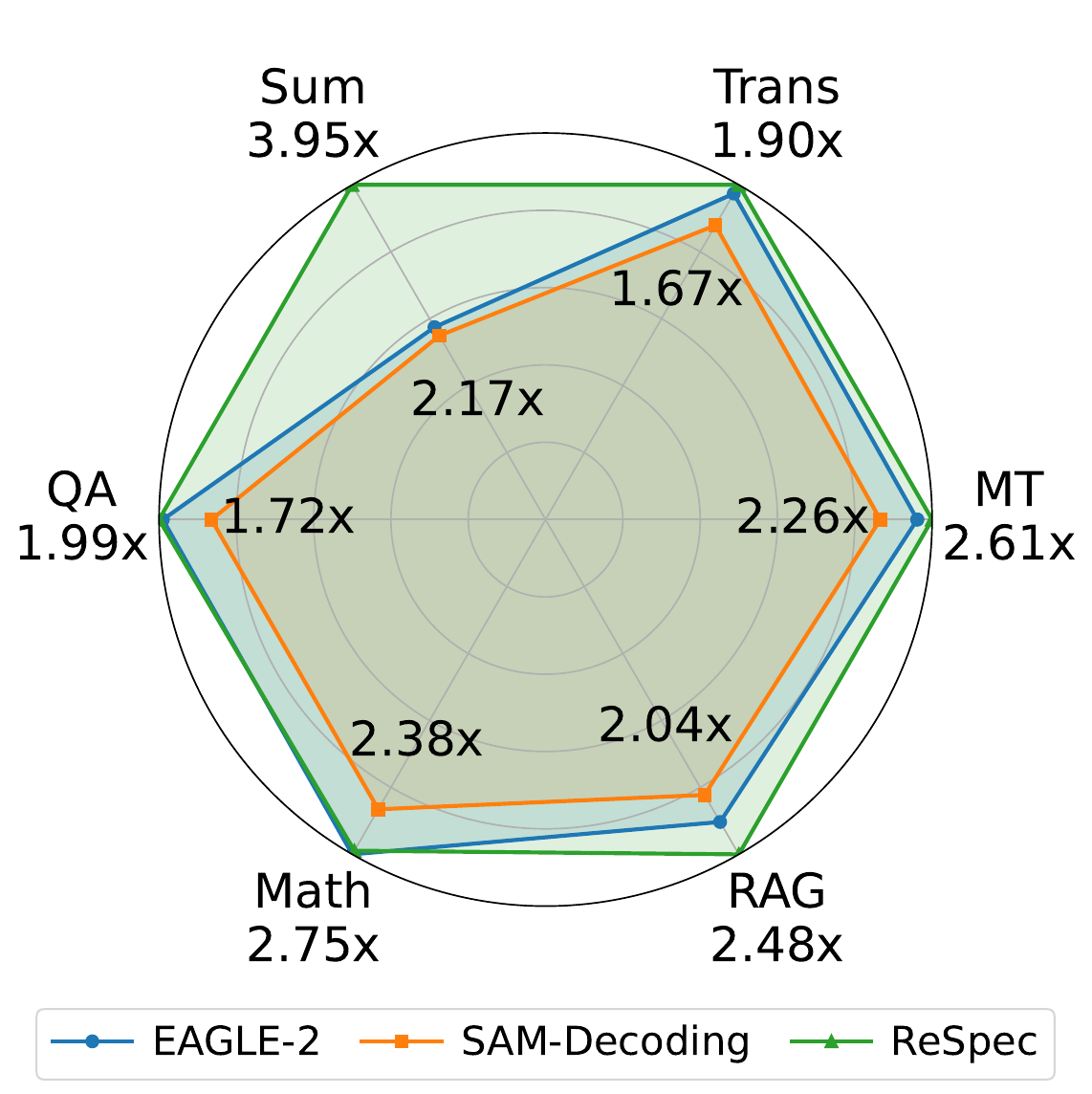}
        \caption{Qwen2-7B-Instruct}
        \label{fig:sub3}
    \end{subfigure}
    \caption{Relative speedup on Spec-Bench tasks. \myFrame~demonstrates strong and balanced performance across all domains, outperforming specialized methods.}
    \label{fig:task_breakdown}
\end{figure}

\textbf{Task-Specific Breakdown.} To provide a more granular analysis of performance, Figure~\ref{fig:task_breakdown} provides a breakdown of speedups across the six tasks in Spec-Bench. The evaluation is conducted on two different base models, Vicuna-7B and Qwen2-7B, to demonstrate the generalizability of \myFrame.

We observe that the advantages of \myFrame~are particularly pronounced in tasks that exhibit high degrees of textual repetition or structural predictability. For example, with the Vicuna-7B model, \myFrame~achieves its peak speedup on Sum (5.21x) and shows strong performance on RAG (2.81x), where the retrieval module provides significant value.
Simultaneously, \myFrame~maintains a robust performance advantage on more reasoning-intensive and less repetitive tasks like Math (2.86x) and QA (2.23x) where the powerful EAGLE-2 component takes over. This highlights the adaptability of our entropy-guided trigger.
A similar pattern is observed with the Qwen2-7B model, where \myFrame~excels in Sum (3.95x) and RAG (2.48x), while still delivering competitive speedups in Math (2.75x) and QA (1.99x) against all baseline.


\subsection{Quality Assurance for Hybrid Verification}
\label{sec:ablation}
To validate the robustness of our hybrid verification strategy, particularly on the Sum task where it achieves significant speedup (Figure~\ref{fig:task_breakdown}), we conduct an ablation study using Qwen2-7B-Instruct model as a representative case. 
We analyze the sensitivity of our method to its core hyperparameters, i.e., the tolerance margin $\tau$ and the minimum look-ahend tolerance greedy match $m$. We analyze the sensitivity of our method to these parameters by measuring their impact on both the mean accept length (MAT) and the final quality, as evaluated by GPT-4o. In the following experiments, EAGLE-2 serves as a performance baseline.

\begin{figure*}[!t]
    \begin{minipage}[t]{0.49\textwidth}
    \centering
        \subcaptionbox{ \#MAT\label{subexp:tmat}}{\includegraphics[width=0.494\textwidth]{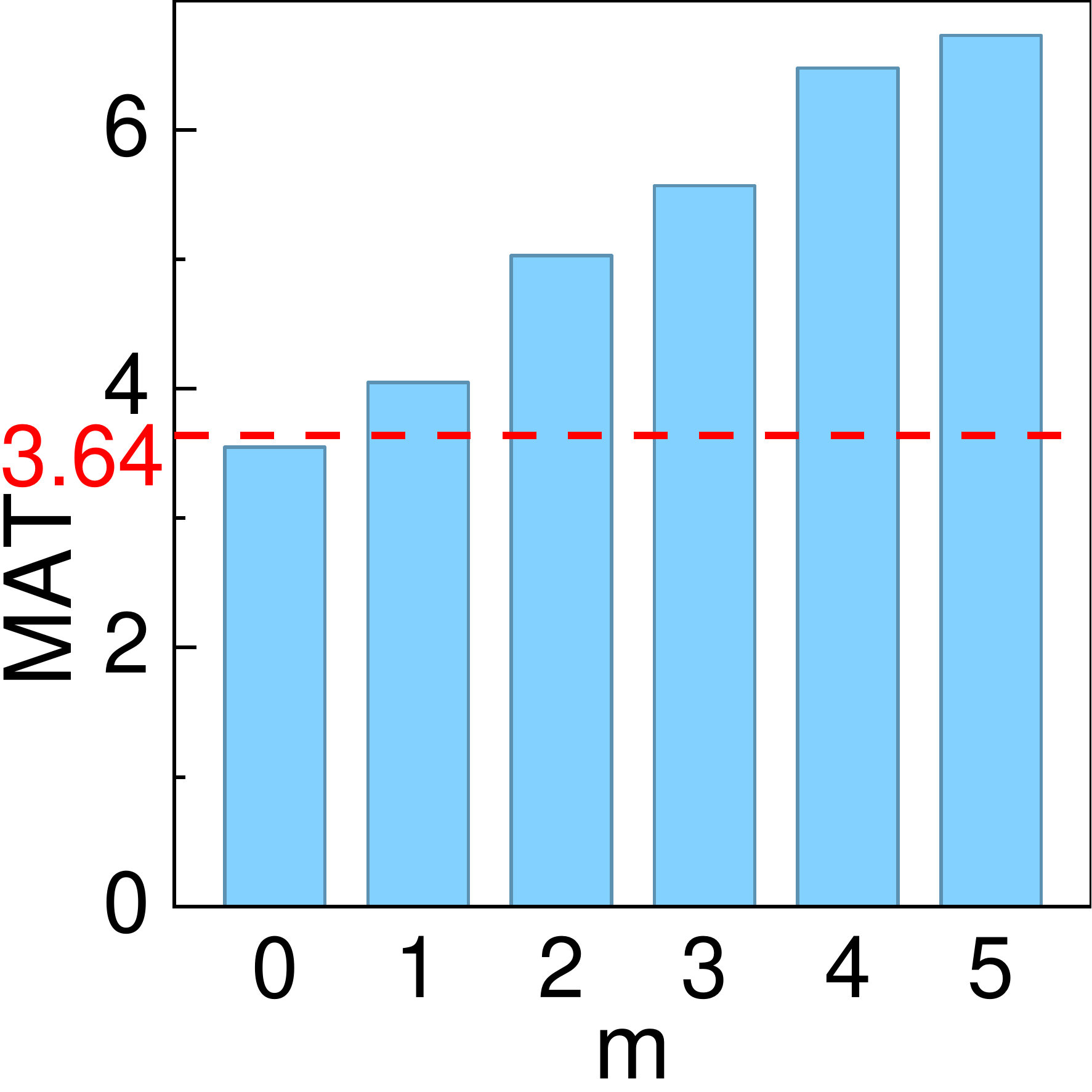}}
        \subcaptionbox{ GPT-4o Score\label{subexp:tscore}}{\includegraphics[width=0.494\textwidth]{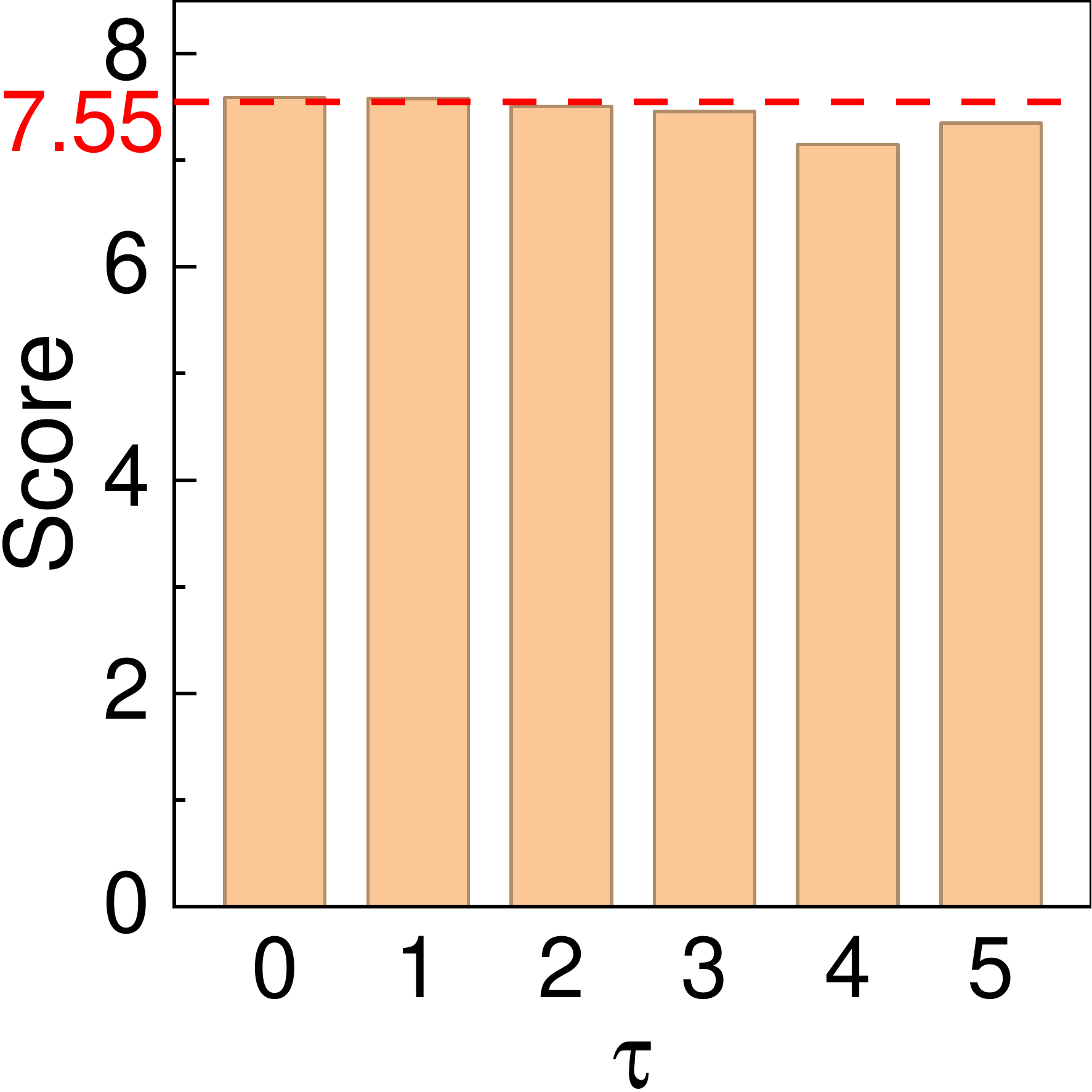}}
        \caption{Effect of $\tau$ on MAT and GPT-4o quality scores for Sum task using Qwen2-7B.}
    \label{fig:ablation_tau}
    \end{minipage}
    \hfill
    \begin{minipage}[t]{0.49\textwidth}
    \centering
        \subcaptionbox{ \#MAT\label{subexp:mmat}}{\includegraphics[width=0.494\textwidth]{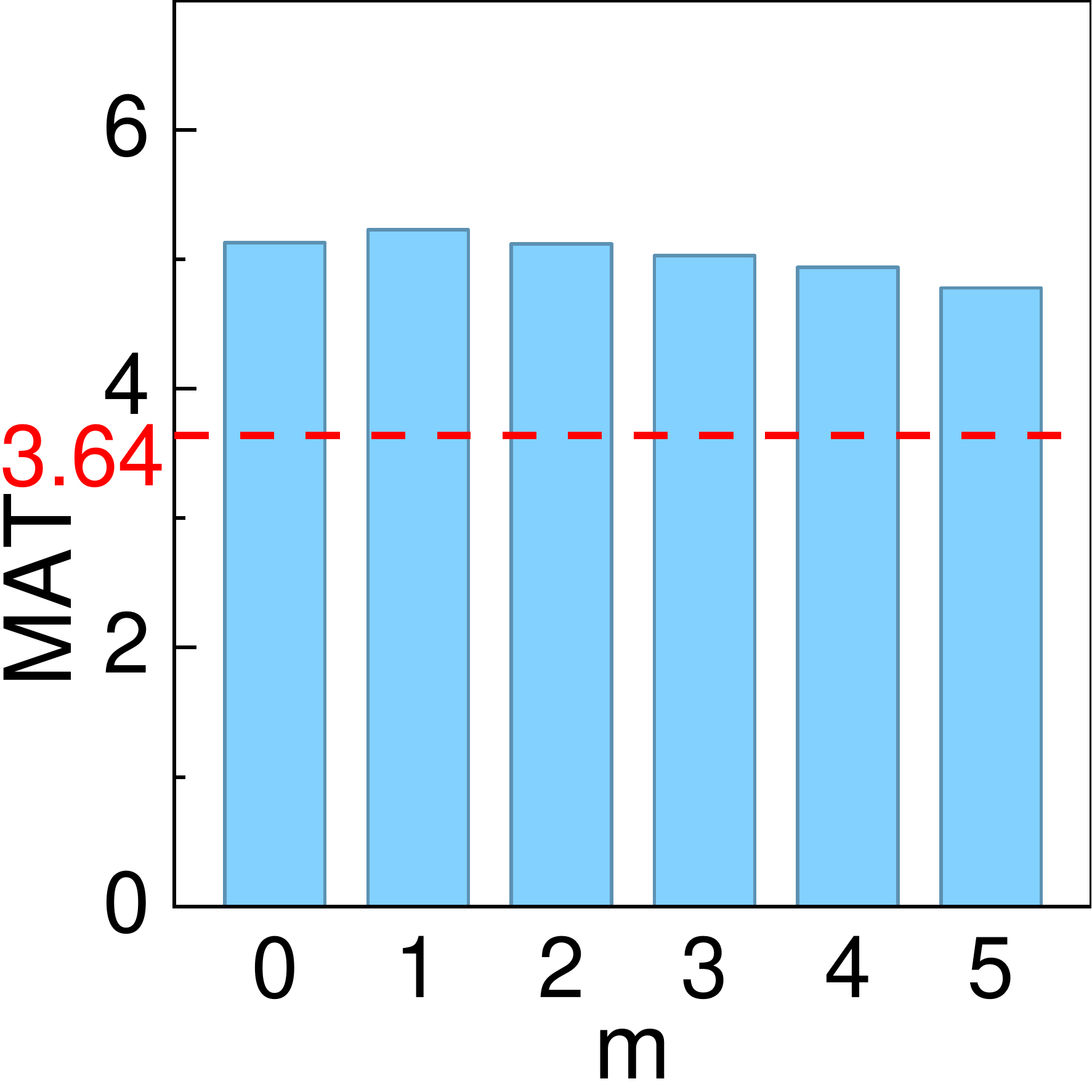}}
        \subcaptionbox{ GPT-4o Score\label{subexp:mscore}}{\includegraphics[width=0.494\textwidth]{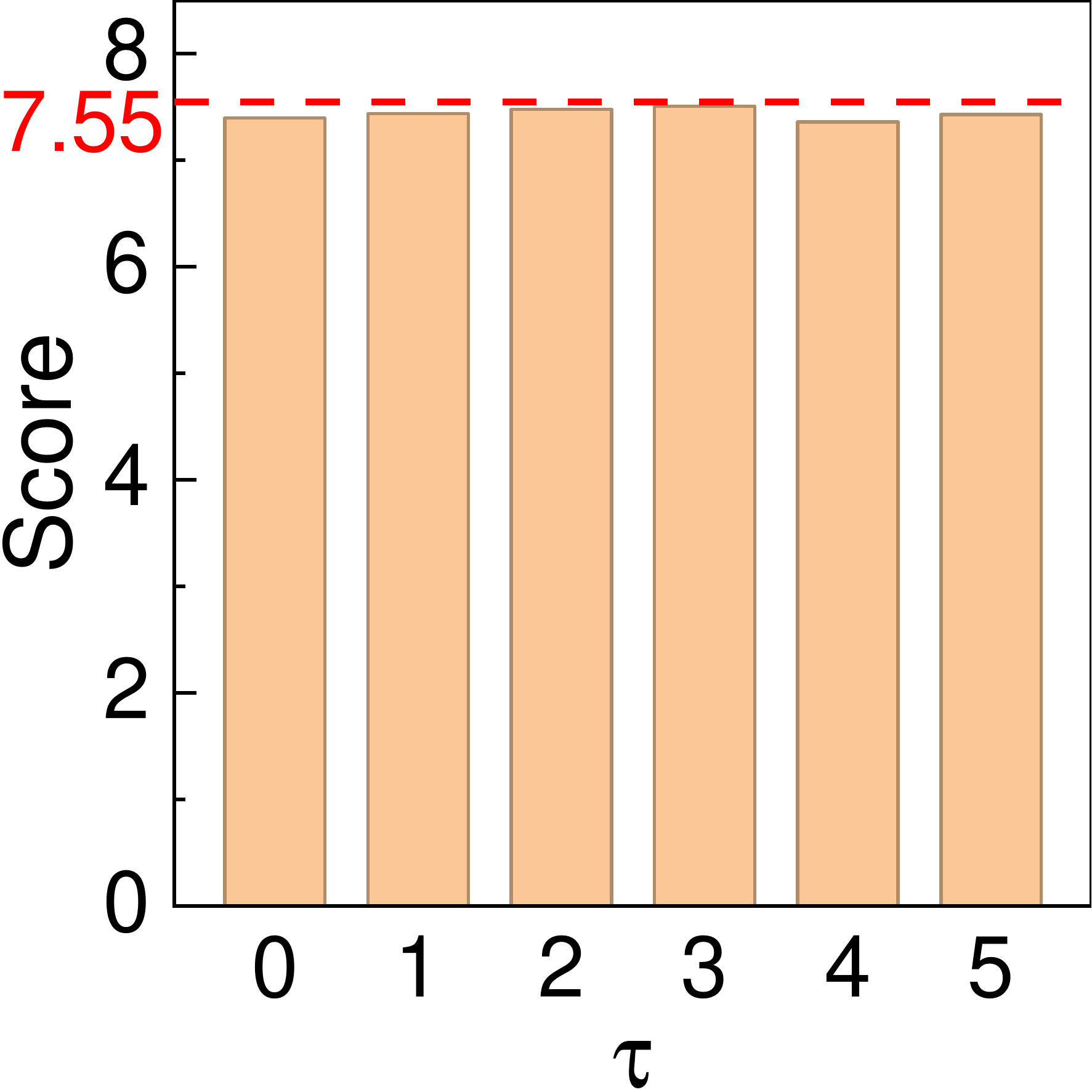}}
        \caption{Effect of $m$ on MAT and GPT-4o quality scores for Sum task using Qwen2-7B.}
    \label{fig:ablation_min_match}
    \end{minipage}
\end{figure*}

\textbf{Sensitivity to tolerance margin $\tau$.} 
We first investigate the system's sensitivity to the leniency parameter $\tau$ in Equation~(\ref{eq:constraint2}), with results presented in Figure~\ref{fig:ablation_tau}. As expected, Figure~\ref{subexp:tmat} shows that MAT is positively correlated with $\tau$. This indicates that a more lenient verification strategy yields a higher token acceptance rate, leading to greater inference acceleration. Crucially, despite the substantial speedup (e.g., nearly doubling the acceptance rate from $\tau=0$ to $\tau=5$), Figure~\ref{subexp:tscore} demonstrates that the final output quality remains robustly high, consistently matching EAGLE-2 baseline. This finding provides strong evidence that $\tau$ can be tuned to significantly boost efficiency without sacrificing quality.

\textbf{Sensitivity to look-ahend tolerance greedy match $m$.}
Similarly, we analyze the impact of the minimum match length, $m$ (Figure~\ref{fig:ablation_min_match}). In contrast to $\tau$, a stricter (higher) $m$ value reduces \#MAT, thereby decreasing the inference speedup, as shown in Figure~\ref{subexp:mmat}. Nevertheless, the GPT-4o quality scores remain stable and competitive with EAGLE-2 baseline across all tested values (Figure~\ref{subexp:mscore}). This illustrates that the quality of our method is not compromised even under more stringent matching conditions, although this comes at the cost of a lower acceptance rate.

Taken together, these studies empirically confirm the robustness of our hybrid verification strategy. The final output quality is largely insensitive to significant variations in its core hyperparameters, reinforcing our central claim that the proposed method achieves substantial efficiency gains without compromising the fidelity of the generated text.
Thus, we recommend setting $\tau \leq 3$ and $m \leq 3$ to maintain a precision error within $1\%$ while retaining significant efficiency gains.

\section{Discussion and Conclusion}
In this work, we introduced \myFrame, a hybrid speculative decoding framework that tackles a core inefficiency in retrieval-enhanced methods, i.e., indiscriminate and costly retrieval attempts. By reframing the copy decision from a heuristic guess to a calculated choice based on contextual uncertainty, our \textbf{entropy-guided adaptive trigger} effectively prunes low-quality retrieval attempts and allocates verification resources where they matter most. Combined with a \textbf{feedback-driven candidate selection} that maximizes the value of each retrieval and a source-aware \textbf{relaxed verification strategy} that pragmatically balances speed and fidelity, pushes the state-of-the-art in speculative decoding. Our experiments demonstrate that \myFrame~achieves significant speedups over powerful baselines like EAGLE-2 and SAM-Decoding without compromising generation quality. The success of \myFrame~underscores the importance of adaptive, context-aware control in hybrid inference systems and points toward a promising direction for future optimization research.


\textbf{Limitations and Future Work:} 
While effective, our method has several limitations. The relaxed verification introduces a minor trade-off between precision and efficiency, which may slightly affect tasks requiring extremely strict alignment. The task-adaptive window length $l$ is currently chosen heuristically, and future work may explore dynamically optimizing this parameter. Furthermore, \myFrame~could be extended by replacing the entropy-based heuristic with a lightweight learned controller, enabling more nuanced and adaptive decision-making across diverse contexts.

In conclusion, \myFrame~offers a more intelligent, efficient, and practical paradigm for accelerating LLM inference. By asking not just \textit{what} to retrieve, but \textit{when} and \textit{how}, we unlock substantial performance gains, paving the way for faster and more responsive large language models.

\bibliography{references}
\bibliographystyle{single-temp/iclr2025_conference}

\end{document}